# Enhanced Directional Smoothing Algorithm for Edge-Preserving Smoothing of Synthetic-Aperture Radar Images

Mario Mastriani and Alberto E. Giraldez

**Abstract**

Synthetic aperture radar (SAR) images are subject to prominent speckle noise, which is generally considered a purely multiplicative noise process. In theory, this multiplicative noise is that the ratio of the standard deviation to the signal value, the "coefficient of variation," is theoretically constant at every point in a SAR image. Most of the filters for speckle reduction are based on this property. Such property is irrelevant for the new filter structure, which is based on directional smoothing (DS) theory, the enhanced directional smoothing (EDS) that removes speckle noise from SAR images without blurring edges. We demonstrate the effectiveness of this new filtering method by comparing it to established speckle noise removal techniques on SAR images.

1. Introduction

Synthetic aperture radar (SAR) imaging of the earth's surface is a valuable modality for remote sensing in Argentina, since SAR is able to penetrate cloud cover and is independent of solar illumination. However, speckle noise generated from the coherent imaging technique of SAR is a serious impediment to computer interpretation of SAR images. This speckle noise can be successfully modeled as a purely multiplicative noise process, and as a result several interesting properties of the noise can be exploited to help reduce the noise without blurring or distorting edges [1]. In theory, the ratio of the standard deviation to the signal value, the "coefficient of variation," is constant at every point in an image corrupted by purely multiplicative noise [1]. This property is not true in all the possible used images [2]. We use a new filter structure independent of such property which is based on directional smoothing (DS) theory [3], the enhanced directional smoothing (EDS) that removes speckle noise from SAR images without blurring edges. The new filter structure is able to direct a filtering operation to act over the complete image. By directing the smoothing operation away from edges, the filter reduces noise while sharpening edges.
Methods used previously to reduce noise in images include speckle filters such as Median, Lee, Kuan, Frost, enhanced Lee, enhanced Frost, Gamma or MAP [4-13], morphology-based nonlinear filter [14], and DS [15,16]. Another possibility is de-noising a SAR image via wavelet shrinkage with a considerable computational complexity [17-24], based on wavelets properties [25-38].

2. Methods

2.1. Speckle Model

Speckle noise in SAR images is usually modeled as a purely multiplicative noise process of the form given in Eq.(1) below. The true radiometric values of the image are represented by $u$, and the values measured by the radar instrument are represented by $v$. The speckle noise is represented by $s$.

$$v(r,c) = u(r,c)\, s(r,c) \tag{1}$$

For single-look SAR images, *s* is Rayleigh distributed (for amplitude images) or negative exponentially distributed (for intensity images) with a mean of 1. For multi-look SAR images with independent looks, *s* has a gamma distribution with a mean of 1. Further details on this noise model are given in [39].

2.2. Speckle Reduction via Enhanced Directional Smoothing

2.2.1 Theory of Enhanced Directional Smoothing

To protect the edges from blurring while smoothing, a directional averaging filter can be useful. Spatial averages *û(r, c:Θ)* are calculated in several directions as

$$\hat{u}(r, c:\Theta) = \frac{1}{N_\Theta} \sum_{k \in W_\Theta} \sum_{l \in W_\Theta} v(r-k, c-l) \tag{2}$$

- which excludes to *v(r, c)* - and a direction *Θ\** is found such that | *v(r, c)* - *û(r, c:Θ\*)* | is minimum. Then

$$\hat{u}(r,c) = \hat{u}(r,c:\Theta^*) \tag{3}$$

gives the desired result for the suitably chosen window *W* and a $N_\Theta$ number of directions, and where *k* and *l* depends on the size of such windows (kernel).
The EDS filter has a speckle reduction approach that performs spatial filtering in a square-moving window know as kernel. The EDS filtering is based on the statistical relationship between the central pixel and its surround-ding pixels as shown in Figure 1.

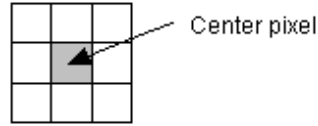

Fig. 1. 3x3 kernel

The typical size of the filter window can range from 3-by-3 to 33-by-33, the size of the window must be odd. A larger filter window means that a larger area of the image can be used for calculation and possibly requires more computation time depending on the complexity of the filter's algorithm. If the size of filter window is too large, the important details will be lost due to over smoothing. On the other hand, if the size of the filter window is too small, speckle reduction may not be very effective. In practice, a 3-by-3 or a 7-by-7 filter window usually yields the best results.

EDS performs the filtering based on either local statistical data given in the filter window to determine the noise variance within the filter window, or estimating the local noise variance using the effective equivalent number of looks (ENL) of a SAR image [24]. The estimated noise variance is then used to determine the amount of smoothing needed for each speckle image. The noise variance determined from the local filter window is more applicable if the intensity of an area is constant or flat whilst ENL is suitable if there are difficulties determining if an area of the image is flat.

### 2.2.2. Algorithms

Algorithm I represents EDS function for four directions and a 3x3 kernel

```
1   function v = eds(v,ROW,COL)
2   for r = 2:ROW-1
3     for c = 2:COL-1
4       d(1) = (v(r,c-1)  +v(r,c+1)  )/2;
5       d(2) = (v(r-1,c)  +v(r+1,c)  )/2;
6       d(3) = (v(r-1,c-1)+v(r+1,c+1))/2;
7       d(4) = (v(r+1,c-1)+v(r-1,c+1))/2;
8       for n = 1:4
9         D(n) = abs(d(n)-v(r,c));
10      end
11      [Dmin,aDmin] = min(D);
12      v(r,c) = d(aDmin);
13    end
14  end
```
Algorithm I

where:
  *v* represents the bitmap matrix of the image
  *eds(•)* is the function that calculate the enhanced directional smoothing of *(•)*
  *ROW* is the number of rows and *COL* is the number of columns of *v*.
  *d* represents the vector of directions
  *D* represents the vector of absolute differences
  *abs(•)* is the function that calculate the absolute value of *(•)*
  *min(•)* is the function that calculate the minimum of vector *(•)* and its location
  *Dmin* is the minimum of vector *D*
  *aDmin* is the location of *Dmin*

Algorithm II represents the homomorphic filter that calls to the **EDS** function.

```
1   [v,map] = imread('namefile.bmp');
2   v = double(v);
3   [ROW,COL] = size(v);
4   v = v + ones(ROW,COL);
5   v = log(v);
6   v = eds(v,ROW,COL);
7   v = exp(v);
8   v = round(v);
9   v = v - ones(ROW,COL);
10  v = uint8(v);
11  imwrite(v,map,'namefileeds.bmp')
```
Algorithm II

where:
  *[v,map] = imread('namefile.bmp');* reads the indexed image in *namefile.bmp* (Windows bitmap) into *v* and its associated colormap into *map*. Colormap values in the image file are automatically rescaled into the range [0,1].
  *double(•)* returns the double precision value for *(•)*
  *size(•)* is the function that calculate the dimensions of matrix *(•)*
  *ones(ROW,COL)* is an ROW-by-COL matrix of ones.
  *log(•)* is the natural logarithm of the elements of *(•)*
  *exp(•)* is the exponential of the elements of *(•)*, e to the *(•)*
  *round(•)* is the function that round towards nearest integer of *(•)*
  *uint8(•)* converts the elements of array *(•)* into unsigned 8-bit integers.
  *imwrite(v,map,namefileeds.bmp)* writes the indexed image in *v*, and its associated colormap *map*, to *namefileeds.bmp*

In Algorithm II, the code line "*v = v + ones(ROW,COL);*" is for avoiding $log(0) = -\infty$.

EDS is applied after chirp scaling algorithm [40] or any algorithm for SAR image generation, so [41].

### 2.2.3. Statistical Measurement

In this work, the assessment parameters that are used to evaluate the performance of speckle reduction are Noise Variance, Mean Square Difference [15,16], Equivalent Number of Looks and Deflection Ratio [24], where:

#### 2.2.3.1. Noise Variance (NV)

NV determines the contents of the speckle in the image. A lower variance gives a "smoother" image as more speckle is reduced, although, it not necessarily depends on the intensity. The formula for calculating the variance is given in Eq.(4)

$$\sigma^2 = \frac{1}{N} \sum_{j=0}^{N-1} u_j^2 \qquad (4)$$

#### 2.2.3.2. Mean Square Difference (MSD)

MSD indicates average difference of the pixels throughout the image where $u_j$ is the denoised image, and $v_j$ is the original image. A higher MSD indicates a greater difference between the original and denoised image. This means that there is a significant speckle reduction. Nevertheless, it is necessary to be very careful with the edges. The formula for the MSD calculation is given in Eq.(5)

$$\text{MSD} = \frac{1}{N} \sum_{j=0}^{N-1} (u_j - v_j)^2 \qquad (5)$$

where N is the size of the image.

#### 2.2.3.3. Equivalent Numbers of Looks (ENL)

Another good approach of estimating the speckle noise level in a SAR image is to measure the ENL over a uniform image region [17]. A larger the value of ENL usually corresponds to a better quantitative performance. The value of ENL also depends on the size of the tested region, theoretically a larger region will produces a higher ENL value than over a smaller region but it also tradeoff the accuracy of the readings. Due to the difficulty in identifying uniform areas in the image, we proposed to divide the image into smaller areas of 25 x 25 pixels, obtain the ENL for each of these smaller areas and finally take the average of these ENL values. The formula for the ENL calculation is given in Eq.(6)

$$\text{ENL} = (\mu/\sigma)^2 \qquad (6)$$

where $\mu$ is the mean of the uniform region and $\sigma$ is the standard deviation of an uniform region The significance of obtaining both MSD and ENL measurements in this work is to analyze the performance of the filter on the overall region as well as in smaller uniform regions.

#### 2.2.3.4. Deflection Ratio (DR)

A third performance estimator that we used in this work is the DR proposed by H. Guo *et al* (1994), [17]. The formula for the deflection calculation is given in Eq.(7)

$$M = (v_{r,c} - v_\mu)/v_\sigma \qquad (7)$$

where $v_{r,c}$ is the scalar pixel value of the image, $v_\mu$ is the estimated mean of $v_{r,c}$ and $v_\sigma$ is the estimated standard deviation of $v_{r,c}$. The ratio M should be higher at pixels with stronger reflector points and lower elsewhere. In H. Guo *et al* 's paper, this ratio is used to measure the performance between different wavelet shrinkage techniques on the diagonal subband only. We instead apply the ratio approach to the same area for wavelet and a kernel that we identify in Fig.1 for our and standard speckle filters.

3. Results

3.1. Performance evaluation

The simulations demonstrate that EDS algorithm improves the speckle reduction performance to the maximum, for single polarization SAR image like fully polarimetric SAR images (when available).

Here, we present a set of experimental results using one ERS SAR Precision Image (PRI) standard of Buenos Aires area. Such image was converted to bitmap file format for its treatment [42].

3.2. Measurements

Fig.2 shows a noisy image used in the experiment from remote sensing satellite ERS-2, with a 540x553 (pixels) x 256 (gray levels) bitmap matrix. Table I summarizes the assessment parameters vs. 11 filters for Fig.2, where En-Lee means Enhanced Lee Filter, and En-Frost means Enhanced Frost Filter. Fig.3 shows the filtered images in the experiment, processed by using eleven speckle reduction schemes: Median, Lee, Kuan, Gamma, Enhanced Lee, Frost, Enhanced Frost, Symlet Wavelets basis 4 and 1 level of decomposition, Daubechies 15 wavelet basis and 1 level of decomposition (improvements were not noticed with other wavelets) [38], DS and EDS filters, respectively.

Fig.3 summarizes the edge preservation performance of EDS vs. the rest of the filters with a considerably smaller computational complexity.

A 3x3 kernel was employed for all statistic speckle filters including EDS.

The assessment parameters NV, MSD, ENL and DR were applied to the whole image.

For Lee, Enhanced Lee, Kuan, Gamma, Frost and Enhanced Frost filters the damping factor is set to 1, see [5-12]. The quantitative results of Table I show that the EDS can eliminate speckle without distorting useful image information and without destroying the important image edges.

In the experiment, EDS outperformed the conventional and no conventional speckle reducing filters in terms of edge preservation measured by Pratt figure of merit [3]. In nearly every case in every homogeneous region, EDS produced the lowest standard deviation and were able to preserve the mean value of the region. The numerical results are further supported by qualitative examination (see Fig. 3).

In the experiment, the filters was applied to complete image, however, only a selected 128x128 pixels windows is showed for image resolutions considerations.

All filters were implemented in MATLAB® (Mathworks, Natick, MA) on a PC with an Athlon (2.4 GHz) processor.

4. Conclusions

In this paper we have developed a new DS algorithmic version based techniques for removing multiplicative noise in SAR imagery. We have shown that with a special filter window (3x3 kernel), the comparison with most commonly used filters (used for SAR imagery [4-38], including wavelets) show lower performance than the EDS for the studied benchmark parameters. This observation has directed us to formulate a new adaptive edge-preserving application of EDS tailored to speckle contaminated imagery. On the other hands, identical results obtained with Symlet wavelet basis 4 and 1 level of decomposition were obtained with the Daubechies wavelet basis 15 and 1 level of decomposition for the experiment.

The EDS exploits the local coefficient of variations in reducing speckle. The performance figures obtained by means of computer simulations reveal that the EDS algorithm provides superior performance in comparison to the above mentioned filters in terms of smoothing uniform regions and preserving edges and features. The effectiveness of the technique encourages the possibility of using the approach in a number of ultrasound and radar applications. Besides, the method is computationally efficient and can significantly reduce the speckle while preserving the resolution of the original image. Considerably increased deflection ratio strongly indicates improvement in detection performance. Also, cleaner images suggest potential improvements for classification and recognition.

References

[1] M. A. Schulze and Q. X. Wu, "Nonlinear Edge-Preserving Smoothing of Synthetic Aperture Radar Images," *Proc. of the New Zealand Image and Vision Computing '95 Workshop*, Christchurch, New Zealand, August 28-29, 65-70 (1995).

[2] P. Dewaele *et al*, "Comparison of some speckle reduction techniques for SAR images". IGARSS, 10:2417-2422, May 1990.

[3] Y. Yu and S. T. Acton, "Speckle Reducing Anisotropic Diffusion", *IEEE Trans. on Image Processing,* vol. 11, no. 11, pp.1260-1270, November 2002.

[4] L. M. Novak *et al*, "Optimal polarimetric processing for enhanced target detection," *IEEE Trans. AES*, 29:234-244, Jan 1993.

[5] R. Fisher *et al*, "Median Filter", http://www.dai.edu.ac.uk/HIPR2/median.htm (Current September 04th, 2001)

[6] J. S. Lee, "Refined filtering of image noise using local statistics," *Comput. Graph.Image Process.,* vol. 15, 380-389, 1981.

[7] J. S. Lee, "Digital Image Enhancement and Noise Filtering by Use of Local Statistics", *IEEE Trans. Pattern Anal. Machine Intell.,* vol. PAMI-2, 1980.

[8] J. S. Lee, "Speckle suppression and analysis for synthetic aperture radar", *Opt. Eng.,* vol. 25, no. 5, pp. 636-643, 1986.

[9] D. T. Kuan *et al*, "Adaptive restoration of images with speckle", *IEEE Trans. Acoust., Speech, Signal Processing,* vol. ASSP-35, pp. 373-383, 1987.

[10] V. S. Frost *et al*, "A model for radar images and its application to adaptive digital filtering of multiplicative noise", *IEEE Trans. Pattern Anal. and Machine Intell.,* vol. PAMI-4, pp. 157-166, 1982.

[11] A. Lopes *et al*, "Adaptive speckle filters and Scene heterogeneity", *IEEE Trans. Geosci. Remote Sensing,* vol. 28, pp. 992-1000, 1990.

[12] A. Lopes *et al*, "Structure detection and statistical adaptive speckle filtering in SAR images", *Int. J. Remote Sensing,* vol. 14, no. 9, pp. 1735-1758, 1993.

[13] Z. Shi and K. B. Fung, "A comparison of Digital Speckle Filters", *Proc. of IGARSS 94,* Canada Centre for Remote Sensing, pp.2129-2133, August 8-12, 1994.

[14] M. A. Schulze and Q. X. Wu, "Noise reduction in synthetic aperture radar imagery using a morphology-based nonlinear filter," *Proceedings of Digital Image Computing: Techniques and Applications,* Conference of the Australian Pattern Recognition Society, Brisbane, Australia, pp.661-666, December 6-8, 1995.

[15] M. Mastriani and A. Giraldez, "Directional Smoothing for Speckle Reduction with Application to Synthetic-Aperture Radar Imagery," submitted to Eurasip Journal on Applied Signal Processing, Special Issue on Advances in Interferometric Synthetic Aperture Radar Processing.

[16] M. Mastriani and A. Giraldez, "Directional Smoothing for Speckle Reduction in Synthetic-Aperture Radar Imagery," submitted to Measurement Science Review.

[17] H. Guo *et al*, "Speckle reduction via wavelet shrinkage with application to SAR based ATD/R", *Technical Report CML TR94-02,* CML, Rice University, Houston, February 1994.

[18] B. B. Hubbard, The World According to Wavelets *(The Story of a Mathematical Technique in the Making)*, A. K. Peter Wellesley, Massachusetts, 1996.

[19] V. R. Melnik *et al*, "A method of speckle removal in one-look SAR images based on Lee filtering and Wavelet denoising", *Proc. of the* IEEE *Nordic Signal Processing Symposium (NORSIG2000),* Kolmarden, Sweden, June 2000.

[20] R. Yu *et al*, "An optimal wavelet thresholding for speckle noise reduction", *In Summer School on Wavelets: Papers,* Publisher: Silesian Technical University (Gliwice, Poland), pp77-81, 1996.

[21] Gao, H.Y., and Bruce, A.G., "WaveShrink with firm shrinkage". *Statistica Sinica*, 7, 855-874, 1997.

[22] L. Gagnon and F. D. Smaili, "Speckle noise reduction of air-borne SAR images with Symmetric Daubechies Wavelets," *SPIE Proc. #2759*, pp. 1424, 1996

[23] L. Gagnon and A. Jouan, "Speckle Filtering of SAR Images - A Comparative Study Between Complex-Wavelet-Based and Standard Filters", *SPIE Proc.#3169, conference "Wavelet Application in Signal and Image Processing V"*, San Diego, 1997.

[24] H. S. Tan, "Denoising of Noise Speckle in Radar Image," http://innovexpo.itee.uq.edu.au/2001/projects/s804298/thesis.pdf

[25] S.G. Mallat, Multiresolution approximations and wavelet orthonormal bases of L2 (R). *Transactions of the American Mathematical Society*, 315(1), pp.69-87, 1989a.

[26] A.Grossman and J.Morlet, "Decomposition of Hardy Functions into Square Integrable Wavelets of Constant Shape," *SIAM J. App Math*, 15: pp.723-736, 1984.

[27] C. Valens, "A really friendly guide to wavelets," http://perso.wanadoo.fr/polyvalens/clemens/wavelets/wavelets.html (current April 12th, 2001)

[28] G. Kaiser, *A Friendly Guide To Wavelets.* Boston: Birkhauser, 1994.

[29] C. S. Burrus *et al*, Introduction to Wavelets and Wavelet Transforms A Primer, Prentice Hall, New Jersey, 1998

[30] I. Daubechies, "Different Perspectives on Wavelets," *Proceedings of Symposia in Applied Mathematics,* vol. 47, American Mathematical Society, United State of America, 1993

[31] J. S. Walker, A Primer on Wavelets and their Scientific Applications, Chapman & Hall/CRC, New York, 1999

[32] E. J. Stollnitz *et al*, *Wavelets for Computer Graphics (Theory and Applications)*, Morgan Kaufmann Publishers, San Francisco, 1996

[33] D. L. Donoho and I.M. Johnstone, Ideal spatial adaptation by wavelet shrinkage. *Biometrika*, 81, 425-455, 1994.

[34] J.Shen and G. Strang, "The zeros of the Daubechies polynomials," *Proc. Amer. Math. Soc.,*1996

[35] D.L. Donoho and I.M. Johnstone, "Adapting to unknown smoothness via wavelet shrinkage," *Journal of the American Statistical Assoc.*, vol. 90, no. 432, pp. 1200-1224, December 1995.

[36] R. R. Coifman and D. L. Donoho, "Translation-invariant de-noising", Antoniadis, A., & Oppenheim, G. (eds), *Lecture Notes in Statistics*, vol. 103. Springer-Verlag, pp 125-150,1995.

[37] M.S Crouse *et al*, "Wavelet-based statistical signal processing using hidden Markov models," *IEEE Trans. Signal Processing*, vol 46, no.4, pp.886-902, April 1998.

[38] M. Misiti *et al*, "Wavelet Toolbox, for use with MATLAB®, User's guide, version 2.1," http://www.rrz.uni-hamburg.de/RRZ/Software/Matlab/Dokumentation/help/pdf_doc/wavelet/wavelet_ug.pdf

[39] J. W. Goodman, "Some fundamental properties of speckle", J. Opt. Soc. Am., 66:1145-1150, November 1976.

[40] R. K. Raney *et al*, "Precision SAR Processing Using Chirp Scaling," *IEEE Trans. Image Processing*, vol. 32, no. 4, pp. 786-799, July 1994.

[41] S. R. DeGraaf, "SAR Imaging via Modern 2-D Spectral Estimation Methods," *IEEE Trans. Image Processing*, vol. 7, no. 5, pp. 729-761, May 1998.

[42] A. K. Jain, "Fundamentals of Digital Image Processing", Englewood Cliffs, NJ, 1989.

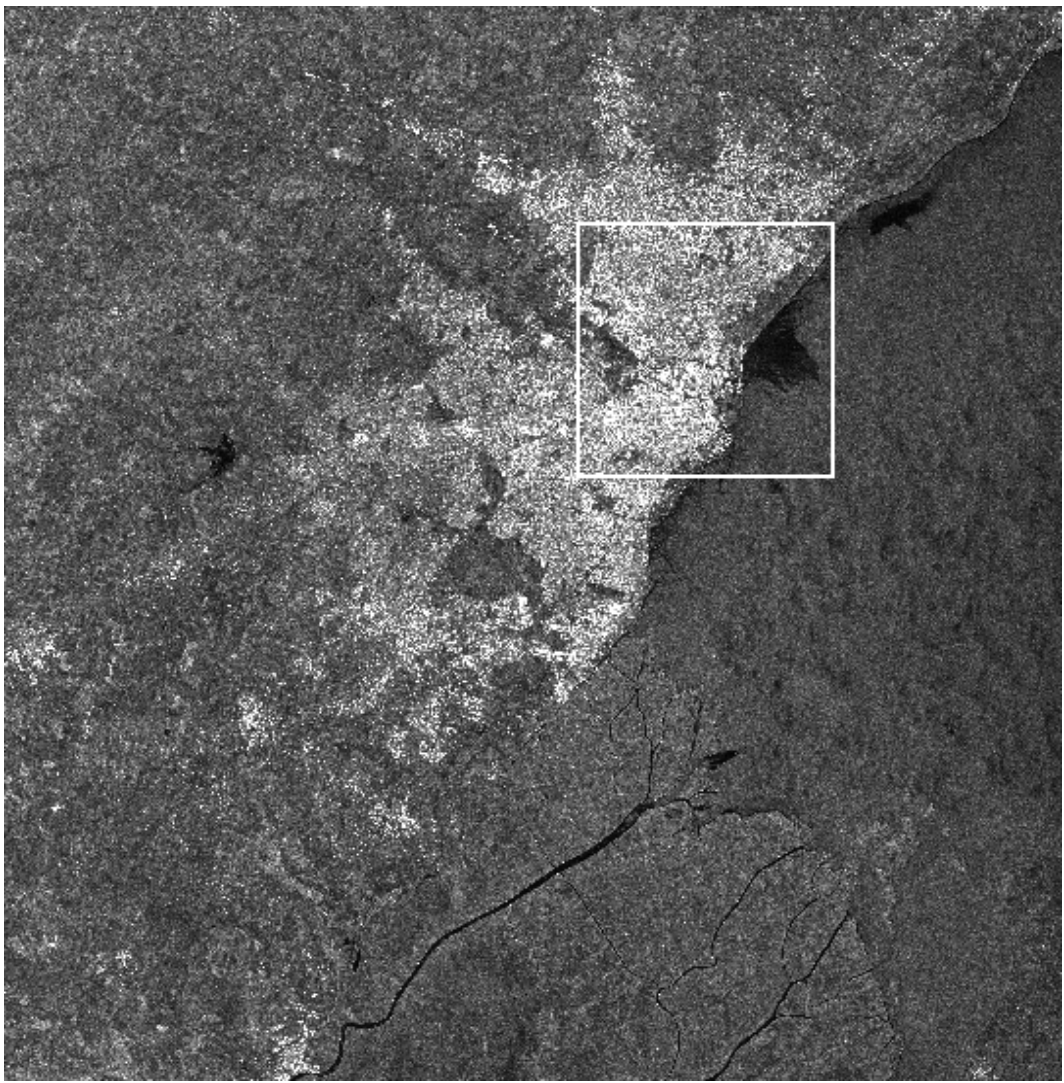

Fig. 2. The white square represents the perimeter of the selected windows (128x128 pixels) of the ERS-2 image for the experiment.

Table I. Assessment Parameters vs. Filters for Fig. 2

| Filter | Assessment Parameters | | | |
|---|---|---|---|---|
| | NV | MSD | ENL | DR |
| Original noisy image | 1.0048e+004 | - | 7.6266 | -4.6797e-004 |
| Wavelet (sym4) | 8.3954e+003 | 888.5013 | 21.3135 | 0.0031 |
| Wavelet (db15) | 8.4049e+003 | 885.6094 | 21.2480 | 3.8729e-005 |
| En-Frost | 7.9526e+003 | 931.0141 | 43.6911 | -0.0014 |
| En-Lee | 7.9521e+003 | 930.8242 | 43.6627 | -0.0014 |
| Frost | 8.1157e+003 | 649.7037 | 36.3795 | -9.4558e-004 |
| Lee | 7.9489e+003 | 939.4810 | 43.9331 | -0.0014 |
| Gamma | 7.9452e+003 | 932.9512 | 43.3836 | -0.0016 |
| Kuan | 8.3265e+003 | 406.7207 | 23.1285 | -0.0013 |
| Median | 7.9524e+003 | 931.3837 | 43.6835 | -0.0014 |
| DS | 8.8840e+003 | 255.1525 | 14.4673 | -0.0011 |
| EDS | 8.5924e+003 | 525.7491 | 19.0819 | -0.0012 |

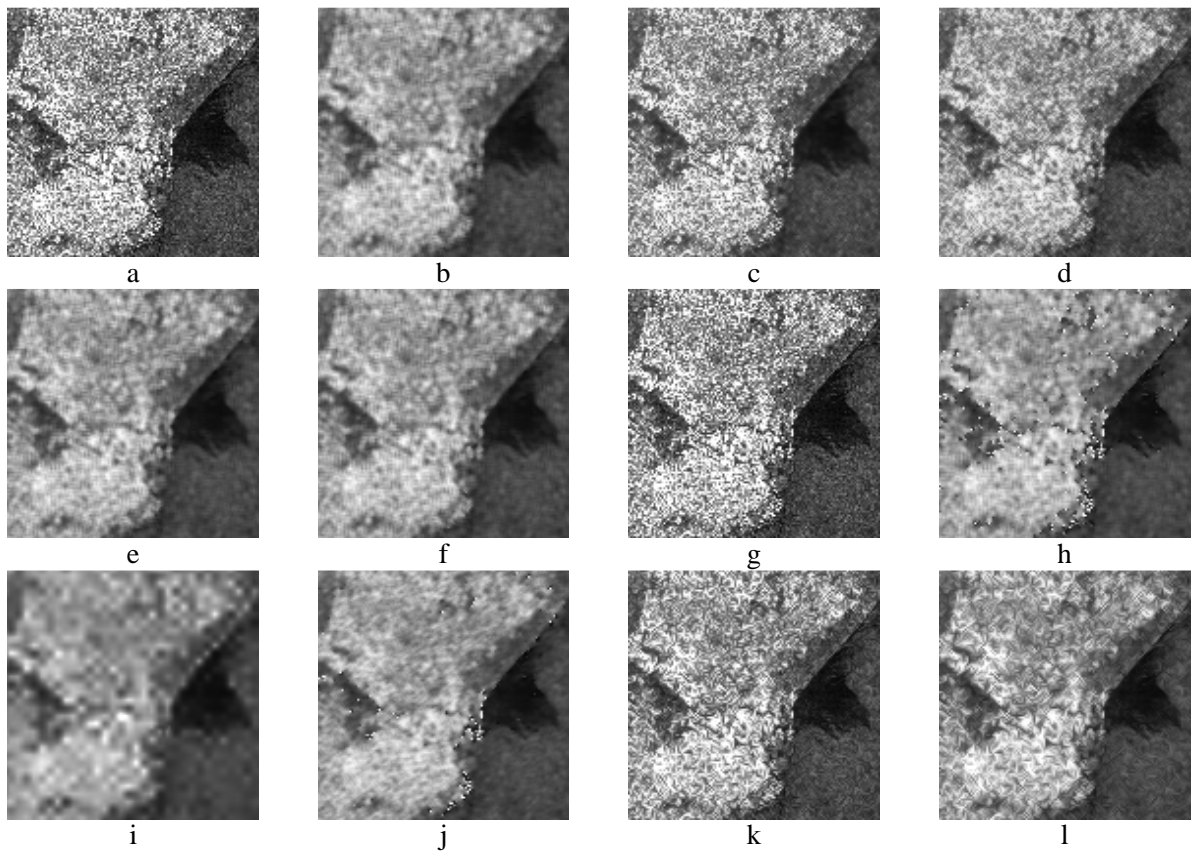

Fig. 3. (a) Original noisy image (Fig. 2). Filtered images from (b) Median, (c) Lee, (d) Kuan, (e) Gamma, (f) Enhanced Lee, (g) Frost, (h) Enhanced Frost, (i) Symlet Wavelets basis 4 and 1 level of decomposition, (j) Daubechies 15 Wavelets basis and 1 level of decomposition, (k) DS and (l) EDS filters.